\documentclass[10pt,twocolumn]{article}

\usepackage[utf8]{inputenc}
\usepackage[T1]{fontenc}
\usepackage[margin=1.8cm, columnsep=0.6cm]{geometry}
\usepackage{times}
\usepackage{microtype}
\usepackage{booktabs}
\usepackage{tabularx}
\usepackage{multirow}
\usepackage{graphicx}
\usepackage{xcolor}
\usepackage{amsmath}
\usepackage{hyperref}
\usepackage{url}
\usepackage{enumitem}
\usepackage{caption}
\usepackage{titlesec}
\usepackage{fancyhdr}
\usepackage{lastpage}
\usepackage{setspace}
\usepackage{dblfloatfix}
\usepackage{balance}

\title{}
\author{}
\date{}

\hypersetup{
    colorlinks  = true,
    linkcolor   = blue,
    citecolor   = blue,
    urlcolor    = blue,
    pdftitle    = {KS-PRET-5M: A 5 Million Word Kashmiri Pretraining Dataset},
    pdfauthor   = {Haq Nawaz Malik, Nahfid Nissar},
    pdfkeywords = {Kashmiri, NLP, dataset, Nastaliq, InPage, Low-Resource}
}

\titleformat{\section}
    {\normalfont\bfseries\normalsize\centering}
    {\thesection.}{0.5em}{\MakeUppercase}

\titleformat{\subsection}
    {\normalfont\bfseries\normalsize}
    {\thesubsection.}{0.5em}{}

\titleformat{\subsubsection}
    {\normalfont\itshape\normalsize}
    {\thesubsubsection.}{0.5em}{}

\titlespacing*{\section}{0pt}{8pt}{4pt}
\titlespacing*{\subsection}{0pt}{6pt}{3pt}

\begin{document}

\twocolumn[{%
\begin{@twocolumnfalse}

\vspace*{6pt}

{\centering
\LARGE\bfseries
KS-PRET-5M: A 5 Million Word, 12 Million Token Kashmiri Pretraining Dataset
\par}

\vspace{16pt}

\begin{center}
\begin{tabular}{cc}

  \begin{tabular}{c}
      \textbf{*Haq Nawaz Malik} \\[3pt]
      \small \href{https://orcid.org/0009-0003-1994-7640}{orcid.org/0009-0003-1994-7640} \\[1pt]
      \small \href{https://huggingface.co/Omarrran}{huggingface.co/Omarrran} \\
      \small \href{https://x.com/HAQ_NAWAZ_MALIK}{x.com/HAQ\_NAWAZ\_MALIK}
  \end{tabular}

  \qquad\qquad

  \begin{tabular}{c}
      \textbf{Nahfid Nissar} \\[3pt]
      \small \href{https://orcid.org/0009-0002-2805-4687}{orcid.org/0009-0002-2805-4687} \\[1pt]
      \small \href{https://huggingface.co/nafiboi}{huggingface.co/nafiboi} \\
      \small \href{https://x.com/NahfidN}{x.com/NahfidN}
  \end{tabular}

\end{tabular}
\end{center}

\vspace{8pt}
\noindent\rule{\textwidth}{0.4pt}
\vspace{4pt}

\noindent{\bfseries\small Abstract.}\quad
{\small
We present \textbf{KS-PRET-5M}, the largest publicly available pretraining
dataset for the Kashmiri language, comprising 5,090,244 (5.09M) words, 27,692,959 (27.6M)
characters, and a vocabulary of 295,433 (295.4K) unique word types. We assembled the
dataset from two source classes: digitized archival and literary material,
encompassing literature, news, biographies, novels, poetry, religious scholarship,
and academic writing, recovered from the proprietary InPage desktop-publishing
format using the converter of Malik~\cite{malik2024inpage}, and Unicode-native
text collected from Kashmiri-language web sources. All text was processed
through an eleven-stage cleaning pipeline that achieves a mean Kashmiri script
ratio of 0.9965, reducing Devanagari contamination to 146 characters across the
full dataset. We tokenized the dataset empirically using
\texttt{google/muril-base-cased}, yielding a subword ratio of 2.383 tokens per
word and a total of approximately 12.13 million subword tokens, substantially
higher than prior estimates derived from non-Kashmiri Perso-Arabic analogues.
KS-PRET-5M is released as a single continuous text stream under CC~BY~4.0
to support language model pretraining, tokenizer training, and computational
linguistic research for Kashmiri.
}

\vspace{6pt}

\vspace{4pt}
\noindent\rule{\textwidth}{0.4pt}
\vspace{10pt}

\end{@twocolumnfalse}
}]

\thispagestyle{fancy}

\section{Introduction}
\label{sec:intro}

Kashmiri is an Indo-Aryan language of the Dardic subgroup with approximately
seven million native speakers, concentrated primarily in the Kashmir Valley.
Despite this speaker population and a literary tradition spanning several
centuries, Kashmiri remains critically underrepresented in the data
infrastructure that underpins modern natural language processing (NLP). Major
multilingual pretraining corpora, including mC4~\cite{xue2021mt5},
CC-100~\cite{conneau2020unsupervised}, and ROOTS~\cite{laurenccon2022bigscience}
contain negligible Kashmiri content, and no standardised evaluation benchmarks
for the language exist. The language's position illustrates a broader pattern
documented by Joshi et al.~\cite{joshi2020state}: the least-resourced languages
receive diminishing returns from the multilingual modelling advances that benefit
better-resourced ones.

A structural obstacle compounds this scarcity. From the 1990s onward, the
InPage desktop publishing system became the dominant tool for professional
Kashmiri typesetting, valued for its high-quality Nastaliq rendering. InPage,
however, uses a proprietary character encoding incompatible with Unicode.
Decades of published Kashmiri literature, journalism, biography, and academic
writing therefore exist in a format invisible to web crawlers and standard
digitization workflows, effectively locking high-quality text out of any
computational pipeline.

Prior to this work, the only available large-scale Kashmiri pretraining resource
was KS-LIT-3M~\cite{malik2026kslit3m}, a 3.1 million word dataset built from
InPage-recovered literary and archival sources. KS-PRET-5M extends this
foundation by incorporating broader genre coverage, including journalism,
biography, academic writing, and religious scholarship, and by adding
Unicode-native web-sourced text, bringing the total to 5.09 million words
and 12.13 million subword tokens.

This paper makes three contributions. First, we demonstrate that large-scale
recovery of professionally typeset Kashmiri text from InPage archives is
achievable at high fidelity, and that this material constitutes the highest-quality
stratum available for Kashmiri NLP. Second, we present and validate an
eleven-stage cleaning pipeline that achieves near-complete script purity across
heterogeneous source material. Third, we provide the first empirical measurement
of the Kashmiri subword tokenization ratio, revealing that widely used
non-Kashmiri multipliers substantially underestimate effective token counts.

\section{Materials and Methods}
\label{sec:methods}

\subsection{Source Material}

KS-PRET-5M draws from two source classes distinguished by their encoding
origin: InPage-recovered archival and literary material, and Unicode-native
web-sourced text.

\subsubsection{InPage-Recovered Archival and Literary Sources}

The highest-quality stratum of Kashmiri text, professionally typeset,
editorially reviewed, and diacritically complete, exists in the InPage format.
InPage does not map cleanly to Unicode: its rendering system interleaves
character codes, positional data, and formatting instructions in a proprietary
structure that generic transliteration tools cannot decode. Earlier conversion
attempts preserved base consonants but systematically lost or misplaced the
diacritical marks that are orthographically obligatory in formal Kashmiri
Nastaliq, rendering the output unsuitable for NLP use.

The converter developed by Malik~\cite{malik2024inpage} addresses this through
reverse-engineered analysis of InPage's internal representation, proper Unicode
combining-character sequencing, and Kashmiri-specific mappings for characters
absent from Urdu and Persian InPage documents. Applying this converter to a
curated collection of InPage files yielded text spanning the following genres:

\begin{itemize}[leftmargin=*,itemsep=2pt,topsep=3pt]
    \item \textbf{Literature and prose fiction:} Short stories, novellas, and
          novels from major Kashmiri presses, contributing stylistically rich
          vocabulary and narrative register.
    \item \textbf{Poetry:} Classical and contemporary Kashmiri verse from
          leading poets of the twentieth and twenty-first centuries,
          providing phonologically precise diacritic usage.
    \item \textbf{News and journalism:} Articles and editorials from
          Kashmiri-language newspapers, contributing contemporary vocabulary,
          named entities, and journalistic register.
    \item \textbf{Biographies and memoirs:} Life writings of Kashmiri public
          figures and scholars, providing biographical vocabulary and
          mixed formal--narrative register.
    \item \textbf{Academic and scholarly writing:} University publications,
          research essays, and academic monographs, contributing technical
          terminology and formal expository register.
    \item \textbf{Religious scholarship:} Commentaries, translations, and
          devotional writings, contributing classical vocabulary and
          traditional textual forms.
    \item \textbf{General prose:} Essays, cultural commentary, and published
          miscellany not captured by the categories above.
\end{itemize}

Source selection applied three criteria: authenticity (professionally published
and editorially reviewed), genre diversity (no single domain contributing a
disproportionate share), and legal compliance .

\subsubsection{Unicode-Native Web Sources}

The second source class consists of Kashmiri-language text already encoded in
Unicode at the time of collection, requiring no format conversion. We collected
content from Unicode-native news portals, cultural blogs, and institutional
websites publishing in Nastaliq. Web sources contribute contemporary language
and current orthographic conventions but introduce higher noise rates,
code-mixing, encoding inconsistencies, and diacritic omission, that the
cleaning pipeline addresses.

\subsection{Eleven-Stage Cleaning Pipeline}

All collected text was processed sequentially through the pipeline summarised
in Table~\ref{tab:pipeline}, implemented in Python~3.12 using \texttt{ftfy},
\texttt{regex}, and \texttt{hashlib}.

\begin{table}[ht]
\caption{Eleven-Stage Cleaning Pipeline Applied to KS-PRET-5M}
\label{tab:pipeline}
\centering
\small
\begin{tabularx}{\columnwidth}{cX}
\toprule
\textbf{Stage} & \textbf{Operation} \\
\midrule
1  & Encoding repair (\texttt{ftfy}): mojibake and broken Unicode \\
2  & Markup removal: Markdown, inline code, HTML/XML tags \\
3  & URL and email address removal \\
4  & Phone number removal (international and local formats) \\
5  & Latin script removal: all English and Roman-script tokens \\
6  & Noise symbol removal: emoji, control characters, box-drawing \\
7  & Western digit string removal: long numeral sequences \\
8  & Whitespace normalisation: multiple spaces and tabs collapsed \\
9  & Line-level filtering: KS ratio, English ratio, digit ratio \\
10 & MD5 exact deduplication \\
11 & Random shuffle (seed\,=\,42); joined into continuous stream \\
\bottomrule
\end{tabularx}
\end{table}

Stage~9 is the primary quality gate. Each line receives a script-purity score
defined as the fraction of non-whitespace characters falling within Arabic and
Nastaliq Unicode blocks (U+0600--U+06FF, U+FB50--U+FDFF, U+FE70--U+FEFF).
Lines scoring below 0.85 are discarded. Table~\ref{tab:unicode} enumerates all
Unicode ranges the pipeline explicitly preserves.

\begin{table}[ht]
\caption{Unicode Ranges Preserved in the Cleaning Pipeline}
\label{tab:unicode}
\centering
\small
\begin{tabularx}{\columnwidth}{lX}
\toprule
\textbf{Range} & \textbf{Content} \\
\midrule
U+0600--U+06FF & Kashmiri Nastaliq (core Arabic block) \\
U+FB50--U+FDFF & Arabic Presentation Forms-A \\
U+FE70--U+FEFF & Arabic Presentation Forms-B \\
U+0750--U+077F & Arabic Supplement \\
U+0900--U+097F & Devanagari Kashmiri \\
U+0660--U+06F9 & Arabic-Indic digits \\
U+060C, U+061B, U+061F, U+06D4 & Kashmiri punctuation \\
Within Arabic block & Diacritics and harakaat \\
\bottomrule
\end{tabularx}
\end{table}

\subsection{Subword Tokenization Measurement}

Token counts for Kashmiri corpora have previously been estimated using a
word-to-subword multiplier derived from non-Kashmiri Perso-Arabic languages
(typically 1.3--1.8). We measure the actual ratio empirically. We loaded
\texttt{google/muril-base-cased}~\cite{khanuja2021muril} via Hugging Face
Transformers and tokenized a 500,000-character sample of the dataset with
truncation disabled. The ratio of subword tokens to whitespace-delimited words
on the sample was then applied to the full dataset. Using a Kashmiri-specific
tokenizer would reduce fragmentation; we chose muril-base-cased as the most
widely adopted multilingual baseline for Indian language research.

\subsection{dataset Format}

The final dataset is distributed as a single continuous UTF-8 text stream.
This format eliminates document-boundary artifacts, enables full context-window
utilisation during pretraining, and integrates directly with Hugging Face
Transformers training pipelines without additional preprocessing.

\section{Results}
\label{sec:results}

\subsection{dataset Scale}

Table~\ref{tab:stats} presents the primary quantitative characteristics of
KS-PRET-5M.

\begin{table}[ht]
\caption{KS-PRET-5M dataset Statistics}
\label{tab:stats}
\centering
\small
\begin{tabularx}{\columnwidth}{lX}
\toprule
\textbf{Metric} & \textbf{Value} \\
\midrule
Total characters          & 27,692,959 \\
Total words               & 5,090,244 \\
Vocabulary size           & 295,433 unique types \\
Type-token ratio (TTR)    & 0.0580 \\
Hapax legomena            & 140,063 \\
Avg.\ word frequency      & 17.23$\times$ \\
Subword tokens (measured) & 12,130,051 \\
Subword ratio             & 2.383 tokens/word \\
Mean KS script ratio      & 0.9965 \\
Devanagari characters     & 146 ($<$0.001\%) \\
Unique Unicode codepoints & 284 \\
File size (UTF-8 TXT)     & 47.88 MB \\
\bottomrule
\end{tabularx}
\end{table}

\subsection{Script Purity}

Table~\ref{tab:purity} presents the character-level script composition.
Nastaliq and Arabic-block characters constitute 81.3\% of the dataset. The
pipeline reduced Devanagari content to 146 characters across 27.7 million
total characters, indicating high-precision separation of Kashmiri text from
mixed-script noise present in the raw web-sourced material.

\begin{table}[ht]
\caption{Character-Level Script Composition of KS-PRET-5M}
\label{tab:purity}
\centering
\small
\begin{tabularx}{\columnwidth}{Xrr}
\toprule
\textbf{Category} & \textbf{Count} & \textbf{Share} \\
\midrule
Nastaliq / Arabic script & 22,510,292 & 81.3\% \\
Kashmiri punctuation     & 156,544    & 0.57\% \\
Arabic-Indic numerals    & 78,626     & 0.28\% \\
Devanagari script        & 146        & $<$0.001\% \\
Whitespace and other     & remainder  & -- \\
\bottomrule
\end{tabularx}
\end{table}

\subsection{Subword Tokenization}

The empirically measured subword ratio of \textbf{2.383 tokens per word}
yields \textbf{12,130,051 subword tokens} for the full dataset. This value
substantially exceeds the 1.3--1.8 range assumed for Perso-Arabic script
languages. The elevated ratio reflects the harakaat density of formal Kashmiri
Nastaliq: each diacritical mark generates one or more additional subword units
under BPE segmentation, because muril-base-cased's multilingual vocabulary does
not contain Kashmiri-specific diacritic-bearing merged forms. A native Kashmiri
BPE tokenizer trained on KS-PRET-5M would learn these merged forms and reduce
the ratio significantly.

\section{Discussion}
\label{sec:discussion}

\subsection{InPage Recovery as a Data Strategy}

The InPage-recovered material forms the highest-quality stratum of KS-PRET-5M.
Because these files were professionally typeset and editorially reviewed before
original publication, they carry complete diacritical marking, consistent
orthography, and grammatically accurate prose, properties that web-sourced
text rarely exhibits. Informal web writing frequently omits harakaat, which is
intelligible to native readers but produces training examples from which a model
cannot learn correct diacritic placement. The breadth of genres recovered,
literature, poetry, journalism, biography, academic writing, and religious
scholarship, ensures that KS-PRET-5M covers registers that no web crawl
would reach.

This outcome demonstrates that the limiting factor for Kashmiri NLP has not
been the absence of text but the absence of tooling to make existing text
computationally accessible. The InPage converter~\cite{malik2024inpage}
effectively unlocks a large archive of professionally produced Kashmiri writing
that predates any Unicode workflow, and further expansion of that archive remains
the most direct path to larger, higher-quality Kashmiri corpora.

\subsection{Tokenization Implications}

The subword ratio of 2.383 has implications beyond KS-PRET-5M. Prior publications
reporting Kashmiri token counts using a 1.3--1.8 multiplier have systematically
understated effective dataset scale, which matters for compute estimation and for
cross-lingual comparisons that use token count as the unit of measurement. The
root cause, BPE fragmentation of harakaat, is well-understood and correctable.
A Kashmiri-native tokenizer trained on this dataset would merge diacritic-bearing
character sequences into single vocabulary entries, reducing the ratio and
improving model efficiency. We encourage future work to measure and report
tokenizer-specific ratios rather than applying language-family estimates.

\subsection{Morphological Coverage}

The hapax legomena rate of 47.4\% (140,063 of 295,433 vocabulary types) is
consistent with expectations for a morphologically complex language at this
dataset scale. Kashmiri exhibits rich verbal inflection, split-ergative agreement,
and nominal case marking, so a single lemma commonly surfaces in dozens of
distinct inflected forms. A high hapax rate at this scale indicates that the
dataset is not dominated by high-frequency repetition; it retains substantial
rare-word coverage necessary for models that must handle the language's full
morphological range.

\subsection{Limitations}

Several limitations constrain the current work. KS-PRET-5M is a pretraining
dataset only; Kashmiri lacks supervised fine-tuning datasets, such as question
answering, natural language inference, and named-entity recognition benchmarks
that are required to evaluate and adapt pretrained models for downstream tasks.
Without such resources, models trained on KS-PRET-5M cannot be rigorously
assessed on task-oriented performance. The dataset also remains limited in
absolute scale: 5 million words is an order of magnitude smaller than the
corpora used for comparable Indian languages such as Hindi and Bengali, and
further expansion across underrepresented domains and dialects is needed. The
dataset reflects the standardised formal written variety of Kashmiri and does
not proportionally represent regional and dialectal variation. Text is provided
without domain labels, speaker metadata, or provenance annotations, which
constrains fine-grained analysis. The subword ratio reported here is
specific to \texttt{muril-base-cased}; researchers using other tokenizers should
measure and report their own ratios.

\section{Conclusions}
\label{sec:conclusions}

KS-PRET-5M is a 5.09 million word, 12.13 million subword token pretraining
dataset for the Kashmiri language, the largest released to date. Three outcomes
warrant particular attention. The recovery of literature, journalism, biography,
and academic writing from InPage archives establishes that professionally
produced Kashmiri text exists at scale and is recoverable at high fidelity; the
InPage encoding barrier, not a shortage of written material, has been the
operative constraint on Kashmiri NLP data. The eleven-stage cleaning pipeline
achieves a mean KS script ratio of 0.9965 with only 146 residual Devanagari
characters, demonstrating that large-scale dataset construction for Kashmiri can
be conducted with high script purity even when source material includes noisy
web text. The empirically measured subword ratio of 2.383 tokens per word
establishes a reproducible tokenization baseline and reveals that prior
dataset-scale estimates for Kashmiri have been systematically understated.

Future work should prioritise three directions: training a Kashmiri-native BPE
tokenizer on KS-PRET-5M to reduce subword fragmentation and improve token
efficiency; continued dataset expansion through further InPage archive recovery
and broader web collection, with explicit attention to dialectal and domain
diversity; and the construction of supervised fine-tuning datasets, covering
tasks such as question answering, named-entity recognition, and textual entailment
to support systematic evaluation of Kashmiri NLP systems.

\section*{Funding Information}

This research received no external funding. The dataset construction was
conducted independently by the authors.

\section*{Conflicts of Interest}

The authors declare no conflicts of interest.

\section*{Data Availability}

KS-PRET-5M is publicly available via the Hugging Face Datasets Hub at:
{\small\url{https://huggingface.co/datasets/Omarrran/KS-PRET-5M_5_million_kashmiri_Pretrainning_LLM_dataset_12M_tokens_2026}}

The dataset is released under CC BY 4.0 with few custom restrictions (Can be seen on Hugging face dataset card info of this dataset).

\section*{Acknowledgments}

The authors thank the Kashmiri literary and journalistic community for
producing and preserving the source material that made this dataset possible.

\bibliographystyle{unsrt}

\begin{thebibliography}{20}

\bibitem{malik2024inpage}
Haq~Nawaz Malik.
\newblock A Robust {InPage}-to-{Unicode} Encoding Converter for {K}ashmiri
  and Related Languages: Addressing a 15-Year-Old Challenge.
\newblock {\em ResearchGate}, 2024.
\newblock \url{https://www.researchgate.net/publication/387522744}.

\bibitem{malik2026kslit3m}
Haq~Nawaz Malik.
\newblock {KS-LIT-3M}: A 3.1 Million Word {K}ashmiri Text Dataset for {LLM}
  Pretraining.
\newblock {\em arXiv preprint arXiv:2601.01091}, 2026.

\bibitem{xue2021mt5}
Linting Xue, Noah Constant, Adam Roberts, Mihir Kale, Rami Al-Rfou, Aditya
  Siddhant, Aditya Barua, and Colin Raffel.
\newblock {mT5}: A Massively Multilingual Pre-trained Text-to-Text Transformer.
\newblock In {\em Proceedings of NAACL 2021}, pages 483--498, 2021.

\bibitem{conneau2020unsupervised}
Alexis Conneau, Kartikay Khandelwal, Naman Goyal, Vishrav Chaudhary, Guillaume
  Wenzek, Francisco Guzm{\'a}n, Edouard Grave, Myle Ott, Luke Zettlemoyer,
  and Veselin Stoyanov.
\newblock Unsupervised Cross-lingual Representation Learning at Scale.
\newblock In {\em Proceedings of ACL 2020}, pages 8440--8451, 2020.

\bibitem{laurenccon2022bigscience}
Hugo Lauren\c{c}on, Lucile Saulnier, Thomas Wang, Christopher Akiki, et~al.
\newblock The {BigScience ROOTS} dataset: A 1.6TB Composite Multilingual
  Dataset.
\newblock In {\em Advances in Neural Information Processing Systems}, 2022.

\bibitem{khanuja2021muril}
Simran Khanuja, Diksha Bansal, Sarvesh Mehtani, Savita Khosa, Atreyee Dey,
  Shachi Dave, Ruchi Garg, Amna Nawaz Khan, and Partha Talukdar.
\newblock {MuRIL}: Multilingual Representations for {I}ndian Languages.
\newblock {\em arXiv preprint arXiv:2103.10730}, 2021.

\bibitem{joshi2020state}
Pratik Joshi, Sebastin Santy, Amar Buber, Kalika Bali, and Monojit Choudhury.
\newblock The State and Fate of Linguistic Diversity and Inclusion in the
  {NLP} World.
\newblock In {\em Proceedings of ACL 2020}, pages 6282--6293, 2020.

\bibitem{kakwani2020indicnlp}
Divyanshu Kakwani, Anoop Kunchukuttan, Satish Golla, Gokul N.C., Avik
  Bhattacharyya, Mitesh~M Khapra, and Pratyush Kumar.
\newblock {IndicNLPSuite}: Monolingual Corpora, Evaluation Benchmarks and
  Pre-trained Multilingual Language Models for {I}ndian Languages.
\newblock In {\em Findings of EMNLP 2020}, pages 4948--4961, 2020.

\bibitem{khan2024sangraha}
Mohammed Safi~Ur~Rahman Khan et~al.
\newblock Sangraha: A Large-Scale Multilingual Dataset for {I}ndian Languages.
\newblock {\em arXiv preprint}, 2024.

\bibitem{bhat2014kashmiri}
Irshad~Ahmad Bhat, Vasu Sharma, Jonathon Read, and Dipti~Misra Sharma.
\newblock A Dependency Treebank of {K}ashmiri.
\newblock In {\em Proceedings of LaTeCH Workshop, ACL 2014}, 2014.

\bibitem{goyal2022flores}
Naman Goyal, Cynthia Gao, Vishrav Chaudhary, Peng-Jen Chen, Guillaume Wenzek,
  Da~Ju, Sanjana Krishnan, Marc'Aurelio Ranzato, Francisco Guzm{\'a}n, and
  Angela Fan.
\newblock The {FLORES-101} Evaluation Benchmark for Low-Resource and
  Multilingual Machine Translation.
\newblock {\em Transactions of the ACL}, 10:522--538, 2022.

\bibitem{devlin2019bert}
Jacob Devlin, Ming-Wei Chang, Kenton Lee, and Kristina Toutanova.
\newblock {BERT}: Pre-training of Deep Bidirectional Transformers for Language
  Understanding.
\newblock In {\em Proceedings of NAACL 2019}, pages 4171--4186, 2019.

\bibitem{conneau2020xlmr}
Alexis Conneau, Kartikay Khandelwal, Naman Goyal, et~al.
\newblock {XLM-R}: Unsupervised Cross-lingual Representation Learning at Scale.
\newblock {\em arXiv preprint arXiv:1911.02116}, 2020.

\end{thebibliography}
\balance

\end{document}